\documentclass[letterpaper]{article} %
\usepackage{aaai2026}  %
\nocopyright
\usepackage{times}                    %
\usepackage{helvet}                   %
\usepackage{courier}                  %
\usepackage[hyphens]{url}             %
\usepackage{graphicx}                 %
\usepackage{natbib}                   %
\usepackage{caption}                  %
\usepackage{graphicx}                 %
\graphicspath{{artifacts/}}
\usepackage{booktabs}  %

\usepackage{booktabs,makecell,array}
\usepackage{booktabs,makecell,array}
\newcolumntype{L}[1]{>{\raggedright\arraybackslash}p{#1}}
\newcolumntype{C}[1]{>{\centering\arraybackslash}p{#1}}

\usepackage{algorithm}
\usepackage{algorithmic}
\usepackage{newfloat}
\usepackage{listings}
\usepackage{amsmath}
\usepackage{booktabs}
\usepackage{tabularx}
\usepackage[hyphens]{url}   %
\DeclareCaptionStyle{ruled}{labelfont=normalfont,labelsep=colon,strut=off}
\floatstyle{ruled}
\newfloat{listing}{tb}{lst}
\floatname{listing}{Listing}

\title{Train-Only–Constrained LM for Synthetic Clinical Trajectories:\\
Leak-Safe Calibration and Auditable Dynamics}

\author{
  Yaser Kord\textsuperscript{\rm 1},
  Moritz Hess\textsuperscript{\rm 2}
}
\affiliations{
  \textsuperscript{\rm 1}Freiburg Center for Data Analysis and Modelling (FDM), University of Freiburg, Germany\\
  \textsuperscript{\rm 2}Institute of Medical Biometry and Statistics, Faculty of Medicine and Medical Center, University of Freiburg, Germany\\
  yaser.kord@fdm.uni-freiburg.de, moritz.hess@uniklinik-freiburg.de
}

\begin{document}
\maketitle

\begin{abstract}
Simulating clinical time series necessitates temporal realism, calibration that preserves model integrity, and privacy transparency. We present a train-only, constrained language-model pipeline that factorizes generation. A compact GPT-style language model learns cross-biomarker co-occurrence through quantile tokens, and temporal directions are enforced by $3{\times}3$ per-biomarker Markov transitions fit on TRAIN, allowing dynamics to be audited. We fit a monotone quantile map (QMap) on TRAIN only (SYN(TRAIN)\,$\rightarrow$\,SIM(TRAIN)); no TEST statistics are used. The fixed map is then applied to SYN(TEST) severities. The constrained LM demonstrated improved distributional fidelity and co-occurrence in a 5,000-patient simulated testbed (parameters within published clinical ranges) compared to an empirical sampler, maintained near-chance detectability after a small $\lambda$-sweep, and exhibits temporal adherence by construction. A classifier-based membership-inference audit results in an AUC of approximately 0.5; we make no differential-privacy claim. We present multi-seed results with paired 95\% confidence intervals and quantify the severity of QMap's alignment ($\Delta W_1$ with confidence interval). This is a Stage-1 method result obtained on a controlled simulator; Stage-2 validation on real EHR cohorts with clinical endpoints is planned.
\end{abstract}

\section{Introduction}
Clinical time series are irregular, privacy-sensitive, and central to healthcare AI. Transformer language models (LMs) on coded EHR sequences (e.g., BEHRT, Med-BERT) learn strong representations \citep{li2020behrt,li2021medbert}, while generators such as TimeGAN and tabular diffusion aim to produce synthetic data \citep{yoon2019timegan,kotelnikov2023tabddpm}. In practice, synthetic clinical generators repeatedly face: (i) \emph{temporal realism}---unconstrained LMs invent implausible visit-to-visit dynamics; (ii) \emph{calibration drift}---weak scales tuned post hoc using test statistics; and (iii) \emph{privacy opacity}---memorization is hard to audit and can be probed via membership-inference attacks (MIA) \citep{shokri2017mia}.

\noindent\textbf{Our approach.} We present a train-only, constrained LM pipeline that factorizes generation. A compact GPT-style model learns cross-biomarker co-occurrence via quantile tokens, while temporal directions ($\downarrow$\,/\,$\rightarrow$\,/\,$\uparrow$; down/flat/up) are enforced by per-biomarker $3{\times}3$ Markov models---an initial-state distribution (INIT) and a transition matrix (TRANS)---fit on TRAIN, making dynamics auditable by construction. For calibration, we adjust only a scalar severity axis (PC1 of TRAIN-standardized quantile features, rescaled to $[0,100]$) with a monotone quantile map (QMap). We denote the ODE-simulated cohort as SIM and LM-generated cohorts as SYN, and fit
\[
f(s) \;=\; Q_{\mathrm{SIM(TRAIN)}}\Big(F_{\mathrm{SYN(TRAIN)}}(s)\Big).
\]
No TEST statistics are used. The fixed $f$ is applied to severities of SYN(TEST); tokens and transitions are unchanged.

\noindent\textbf{Validation strategy (Stage 1).} We evaluate on an ODE-driven simulator of longitudinal biomarkers ($\approx$5,000 synthetic patients; parameters within clinical ranges). Patients are split 70/15/15 into TRAIN/VAL/TEST. Because the data-generating dynamics are known, we can test---falsifiably---whether the model preserves both transition behavior and cross-biomarker structure. This is a first-stage methods study; evaluation on real EHR cohorts with clinical endpoints is planned for Stage 2.

\noindent\textbf{Contributions.}
\begin{itemize}\setlength{\itemsep}{2pt}
\item \textbf{Factorized generator.} Within-visit co-occurrence is modeled by a compact LM; per-biomarker $3\times3$ Markov chains (INIT/TRANS) govern visit-to-visit directions, fit on TRAIN.
\item \textbf{Leak-safe calibration.} A monotone quantile map is fit using SYN(TRAIN)\,$\rightarrow$\,SIM(TRAIN) and then applied to SYN(TEST); tokens/transitions are unchanged and no TEST statistics are used.
\item \textbf{Audits with uncertainty.} We report paired 95\% CIs for distributional match (MMD$^2$, Energy, Detect-AUC), co-occurrence ($\Sigma$-gap), temporal agreement (L1, $\chi^2$), and a classifier-based membership-inference (MIA) AUROC.
\item \textbf{Lightweight and reproducible.} Small footprint; priors, scalers, and run manifests are persisted to enable exact reruns and auditing.
\end{itemize}

\section{Method}
\label{sec:method}
\textbf{Data and split (SIM).} We use an ODE-simulated cohort SIM (six biomarkers; irregular visits), split 70/15/15 into TRAIN/VAL/TEST (e.g., $N\approx5000$). All decoding and calibration statistics (cutpoints, INIT/TRANS, scaler, PCA, QMap) are fit on SIM(TRAIN) only; VAL selects early stopping and $\lambda$; TEST is held out.

\noindent\textbf{Overview.} We generate visit-level token sequences for $B$ biomarkers. A compact decoder-only LM learns content co-occurrence; a per-biomarker 3-state Markov prior (DOWN/FLAT/UP) enforces temporal directions at decode time; and a TRAIN-only monotone quantile map aligns a 1D severity axis between synthetic (SYN) and real (SIM) data. Unless stated otherwise, all scalers, PCA, and maps are fit on SIM(TRAIN). We use the shorthand SIM(TRAIN/TEST) and SYN(TRAIN/TEST) throughout.

\subsection{Motivation}
Synthetic EHR generators face three recurring issues: (i) temporal drift when unconstrained LMs hallucinate jumps; (ii) calibration leakage when clinical scales are aligned using test data; and (iii) privacy opacity from black-box generators. Our pipeline addresses these with constrained decoding (limits drift), TRAIN-only monotone quantile mapping (prevents leakage), and explicit Markov transitions (auditable dynamics).

\subsection{Mechanistic prior and tokens}
A clinically motivated biomarker ODE guides which aspects to constrain. Each visit emits, for each biomarker, a level token \texttt{q1..q5} (marginal level) and a direction token \texttt{dir:\{DOWN,FLAT,UP\}} (local trend), separated by \texttt{<STEP>}. A canonicalizer fills any missing per-visit token. This discrete design (i) keeps the state space small and interpretable, (ii) separates \emph{what} changes (level) from \emph{how} it changes (direction), and (iii) makes dynamics easy to audit. \textit{Scalability.} Vocabulary grows linearly with biomarkers $B$: per visit we emit $B$ level tokens and $B$ direction tokens (six biomarkers in this paper). All constraints (TRAIN-fit marginals and INIT/TRANS), the severity axis, and audits apply identically for any $B$. \textit{Training scope.} All priors and statistics referenced below are estimated on SIM(TRAIN) only.

\subsection{Compact, decoder-only LM}
We train a small GPT-style decoder on patient sequences \texttt{<BOS>}\ldots\texttt{<EOS>} with cross-entropy (AdamW, early stopping). The LM captures content co-occurrence and visit rhythm; \emph{temporal directions are not learned} and are imposed only at decoding. We fix seeds, use deterministic kernels, and record a run manifest (configs, hashes, metrics) for exact re-runs. Full hyperparameters are in App.~A.

\subsection{Constrained decoding}
\textbf{Levels (quantiles).} We combine LM evidence with TRAIN marginals:
\[
\tilde{\ell}_q=\ell^{\mathrm{LM}}_q+\lambda\log p_{\mathrm{TRAIN}}(q),\qquad q\in\{\mathrm{q1..q5}\},
\]
then sample with temperature/top-$p$ (details in App.~A). This preserves co-occurrence while discouraging out-of-support levels. We sweep $\lambda\in\{0.3,0.5,0.7\}$ on VAL and pick a near-chance detectability operating point.

\noindent\textbf{Directions (UP/FLAT/DOWN).} For each biomarker we sample from a Laplace-smoothed Markov prior fit on SIM(TRAIN): $d_1\sim\mathrm{INIT}$, $d_t\sim\mathrm{TRANS}(d_{t-1},\cdot)$, $t>1$, with $\alpha{=}1$ smoothing. With $N\approx$5000 patients, transition estimates are stable; we keep Laplace smoothing $\alpha{=}1$ to handle rare states. Transition matrices are $3{\times}3$, per biomarker, and independent of quantile bin count. These constraints prevent LM-invented jumps and faithfully reproduce TRAIN flip/persistence rates; they do not ``correct'' implausible patterns present in TRAIN.

\subsection{Train-Fit Severity \& Leak-Safe Calibration}
\label{subsec:severity-calibration}
\textbf{Addressing calibration leakage.} Severity is defined as PC1 from a PCA fit on SIM(TRAIN) standardized quantile features and rescaled to $[0,100]$. We fit a monotone, rank-preserving quantile map on TRAIN only,
\[
f(s) \;=\; Q_{\mathrm{SIM(TRAIN)}}\Big(F_{\mathrm{SYN(TRAIN)}}(s)\Big),
\]
and apply the fixed $f$ to SYN(TEST) severities. Tokens and transitions are unchanged; no TEST statistics are used.

\subsection{Metrics, audits, and uncertainty}
\label{subsec:metrics}
\textbf{Setup.} Unless noted, all scalers, PCA, and $f$ (QMap) are fit on SIM(TRAIN) only. For severity figures/tables we report results before/after TRAIN-calibrated $f$.
\begin{itemize}\setlength{\itemsep}{2pt}
\item \emph{Distribution fidelity:} RBF MMD$^2$ and Energy compare SYN(TEST) vs.\ SIM(TEST); detectability via PCA$\rightarrow$logistic AUC (scaler/PCA fit on SIM(TRAIN)).
\item \emph{Temporal fidelity:} per-biomarker transition $\chi^2$/$p$ and L1 verify UP/FLAT/DOWN dynamics; $p$-values are descriptive. We also report per-biomarker 1-Wasserstein distances over level tokens (units: quantile steps).
\item \emph{Co-occurrence:} covariance heatmaps (SIM vs.\ SYN) and $\Sigma$-gap $=\|\mathrm{Cov}_{\mathrm{SIM}}-\mathrm{Cov}_{\mathrm{SYN}}\|_F$ with paired 95\% bootstrap CIs.
\item \emph{Privacy (descriptive):} (i) TRAIN-only 1-NN audit with a 5th-percentile leave-one-out baseline, and (ii) a PCA$\rightarrow$logistic discriminator reporting AUROC with 95\% CIs. No differential-privacy claim.
\item \emph{Utility:} ridge models trained on SIM(TRAIN) vs.\ SYN(TRAIN) (and optionally SIM$\cup$SYN) and tested on SIM(TEST); report R$^2$/MAE with paired CIs.
\end{itemize}
\textbf{Uncertainty and robustness.} We average over $S{=}5$ random seeds for headline metrics (Detect AUC, MMD$^2$, Energy, Utility) and report mean$\pm$95\% CI. We also use patient-level paired bootstraps ($B{=}2000$) for all metrics and SIM--SYN deltas, reporting percentile 95\% CIs. Unless noted, $\Delta\equiv\mathrm{SIM}-\mathrm{SYN}$; for $\uparrow$ metrics positive $\Delta$ favors SIM, for $\downarrow$ metrics negative $\Delta$ favors SIM. Tables label seed vs.\ bootstrap CIs explicitly.

\begin{table}[t]
  \centering
  \footnotesize                                   %
  \setlength{\tabcolsep}{5pt}                     %
  \renewcommand{\arraystretch}{1.05}              %
  \begin{tabular*}{\columnwidth}{@{\extracolsep{\fill}}lcc@{}}  %
    \toprule
    Policy & Tests saved (\%) & Missed-event rate (\%)\\
    \midrule
    Baseline (measure-all) & 0.0 [0.0, 0.0] & 0.0 [0.0, 0.0]\\
    Threshold 0.3 & 15.7 [12.7, 18.7] & 16.0 [9.4, 23.6]\\
    Threshold 0.5 & 49.2 [45.0, 53.2] & 51.1 [42.1, 60.1]\\
    Threshold 0.7 & 87.2 [84.7, 89.7] & 85.2 [78.1, 91.3]\\
    \bottomrule
  \end{tabular*}
  \caption{Utility vignette: tests saved vs.\ missed-event rate (95\% CIs).}
  \label{tab:policy_vignette}
\end{table}

\subsubsection*{Privacy Audit (TRAIN-only, 1-NN and classifier-MIA)}
We additionally report a PCA$\rightarrow$logistic classifier MIA AUROC with 95\% CIs. For the 1-NN audit, distances are computed in a feature space standardized by a \emph{StandardScaler fit on SIM(TRAIN) only}. As a conservative baseline, we compute leave-one-out 2-NN distances among SIM(TRAIN) patients and define the leak threshold as the baseline 5th percentile. We then compute 1-NN distances from SYN(TEST) and SIM(TEST) to SIM(TRAIN) and report the proportion below threshold (\%\,$<$\,thr) with 95\% CIs (Wilson or Clopper--Pearson). With $N\approx$5000 patients we obtain tighter patient-level paired bootstrap CIs; transition-test $p$-values are descriptive, not confirmatory.

\section{Experimental Setup}
\textbf{Summary.} We train a small LM on tokenized visits and, at decode time, enforce TRAIN-only priors (marginals + Markov directions); a TRAIN-fit PC1 + monotone QMap calibrates severities; we evaluate distribution/temporal/privacy/utility on TEST.

\textbf{Cohort and split.}
We study a 5,000-patient cohort with irregular, clinician-triggered visits. To avoid any patient-level leakage, we perform a \textbf{patient-wise 70/15/15} split into TRAIN/VAL/TEST. All statistics used for decoding constraints and calibration (token quantiles, INIT/TRANS, PCA axis, QMap) are \emph{fit on TRAIN only}. VAL is used for early stopping and sampling knobs; TEST is reserved for the final hold-out evaluation. We persist the split manifest and token vocabulary for exact reruns. With $N{\approx}5000$ patients in TRAIN, the INIT/TRANS matrices are estimated with sufficient support for stable decoding, while Laplace smoothing ($\alpha=1$) mitigates rare-state sparsity.

\textbf{Preprocessing and tokenization.}
Each visit provides \textbf{6 biomarkers} (readily generalizable to $B$ biomarkers). For each biomarker we emit (i) a level token in \texttt{q1..q5} (i.e., \emph{five} quantiles) using TRAIN empirical cutpoints and (ii) a trend token \texttt{dir:\{DOWN,FLAT,UP\}} computed from successive visits with a small min-gap to suppress jitter; any \texttt{START} is coerced to \texttt{FLAT}. Tokens for the six biomarkers are concatenated in a fixed order and visits are separated by \texttt{<STEP>}. A canonicalization pass repairs missing per-visit tokens (carry-forward for direction, nearest quantile for level) to maintain a valid sequence grammar.

\noindent\emph{Example tokens.} \textbf{REAL(TRAIN)} visit:
\texttt{b1:q3 b2:q2 b3:q4 b4:q3 b5:q2 b6:q3\; b1:dir:UP b2:dir:FLAT b3:dir:DOWN b4:dir:FLAT b5:dir:UP b6:dir:FLAT}.
A \textbf{SYN(TRAIN)} sample under the same grammar might be:
\texttt{b1:q3 b2:q3 b3:q4 b4:q2 b5:q2 b6:q3\; b1:dir:FLAT b2:dir:UP b3:dir:FLAT b4:dir:FLAT b5:dir:DOWN b6:dir:FLAT}.
\textbf{REAL(TEST)} is held out and never used for fitting; visits are concatenated with \texttt{<STEP>}.

\textbf{Mechanistic prior.}
A biomarker ODE motivates (i) discretizing values as quantiles (robust across scales) and (ii) constraining temporal directions by a \emph{train-fit} Markov model during decoding. The prior is used to guide design rather than to generate labels on TEST; nothing is tuned on REAL(TEST).

\textbf{Model.}
We train a compact decoder-only Transformer (dim=192, layers=2, heads=3, dropout=0.2) with cross-entropy (AdamW, early stopping on VAL). We fix seeds and batch order and log configuration, file hashes, and metrics to a manifest for reproducibility.

\textbf{Decoding constraints (sampling time).}
For \emph{quantile} tokens we blend LM logits with TRAIN log-marginals
$\tilde{\ell}=\ell_{\mathrm{LM}}+\lambda\log p_{\mathrm{TRAIN}}(q)$
and sample with temperature/top-$p$ (defaults: $\lambda=0.5$, $\tau=1.0$, top-$p=0.98$).
For \emph{direction} tokens $d_t\in\{\downarrow,\rightarrow,\uparrow\}$ we \emph{do not} use LM logits; we draw from TRAIN initial/transition matrices with Laplace smoothing ($\alpha=1$): $d_1\!\sim\!\mathrm{INIT}_{\mathrm{TRAIN}}$, $d_t\!\sim\!\mathrm{TRANS}_{\mathrm{TRAIN}}(d_{t-1},\cdot)$. This yields auditable, train-fit temporal dynamics by construction.

\textbf{Severity axis and calibration.}
From REAL(TRAIN) quantile features we fit a 1-D PCA axis (PC1) and rescale to $[0,100]$; this axis remains fixed. We then learn a \emph{monotone} quantile map on TRAIN only,
$\mathcal{Q}(x)=F^{-1}_{\text{real-train}}\!\big(F_{\text{syn-train}}(x)\big)$,
and apply it to SYN(TEST) severities, avoiding any use of REAL(TEST) during fitting \citep{chernozhukov2017generic}.

\textbf{Baselines and ablations.}
(1) \textbf{Unconstrained LM}: same architecture, no prior blending; directions from LM logits (no Markov constraints). 
(2) \textbf{QMap Before vs After}: alignment ablation on the fixed TRAIN-fit severity axis.

\textbf{Metrics and uncertainty.}
(1) \emph{Distribution fidelity}: RBF MMD$^2$ (median-heuristic bandwidth) and Energy; detectability via PCA+logistic AUC. 
(2) \emph{Temporal fidelity}: $\chi^2$/$p$ and L1 on direction transition matrices (aggregate and per biomarker). 
(3) \emph{Privacy}: 1-NN distances of SYN(TEST)$\to$REAL(TRAIN), compared with the conservative REAL(TRAIN)$\to$REAL(TRAIN) 5th-percentile self-baseline; this is contextualized by membership-inference risk literature without claiming DP guarantees \citep{shokri2017mia,abadi2016dpsgd}. 
(4) \emph{Utility}: Ridge (regression) and Logistic (classification) trained on \textbf{REAL(TRAIN)} vs \textbf{SYN(TRAIN)} and evaluated on \textbf{REAL(TEST)}; regularization is selected using TRAIN/VAL only.
We report \emph{patient-level paired bootstrap} CIs (B=2000). With small-$N$, transition-level $p$-values are treated as descriptive rather than confirmatory.

\section{Results}

\subsection{Distribution, Temporal, and Privacy Audits}

\begin{figure}[t]
\centering
\begin{minipage}[t]{0.49\columnwidth}
  \centering
  \includegraphics[width=\linewidth]{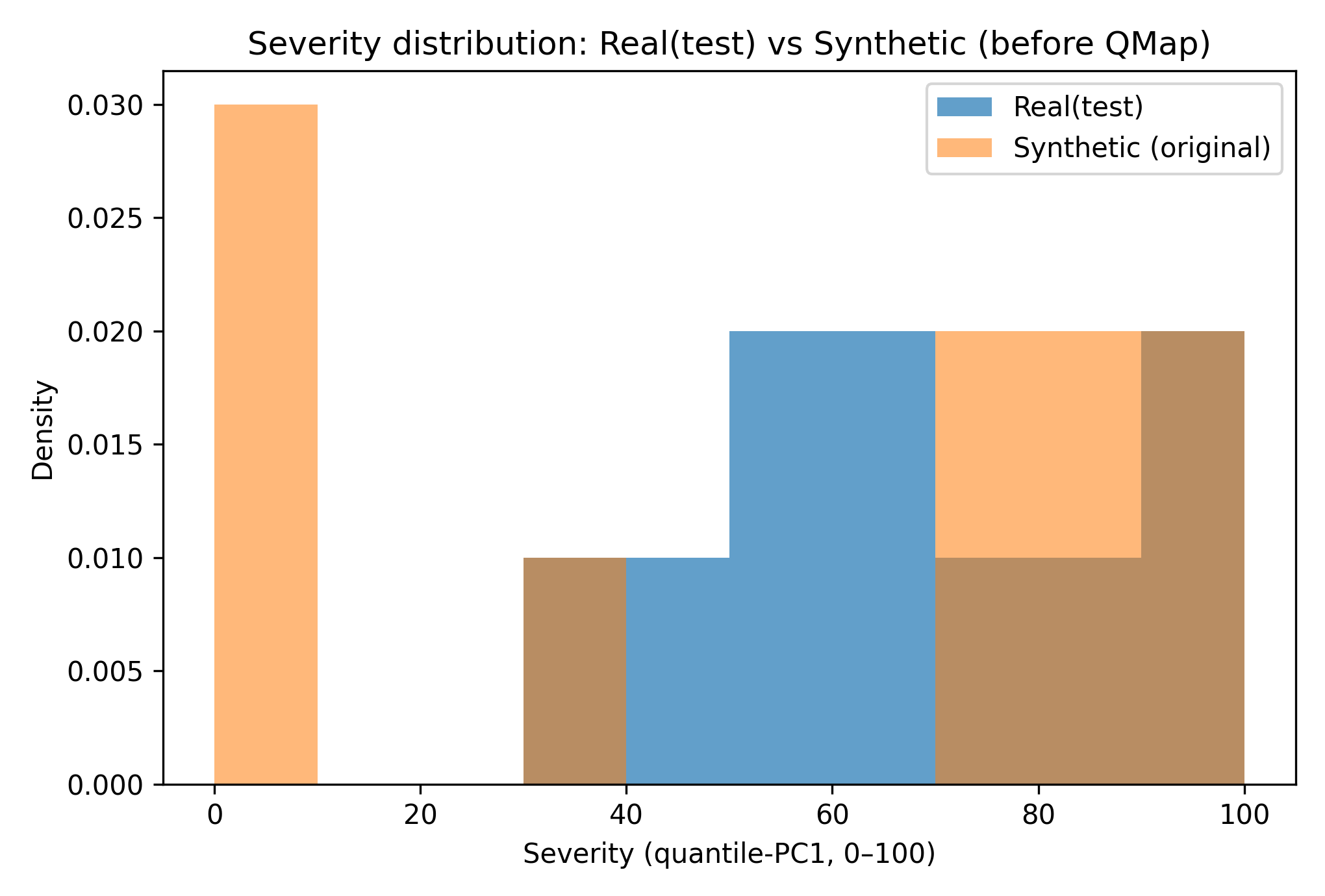}
  \vspace{1mm}\footnotesize Before QMap\par
\end{minipage}\hfill
\begin{minipage}[t]{0.49\columnwidth}
  \centering
  \includegraphics[width=\linewidth]{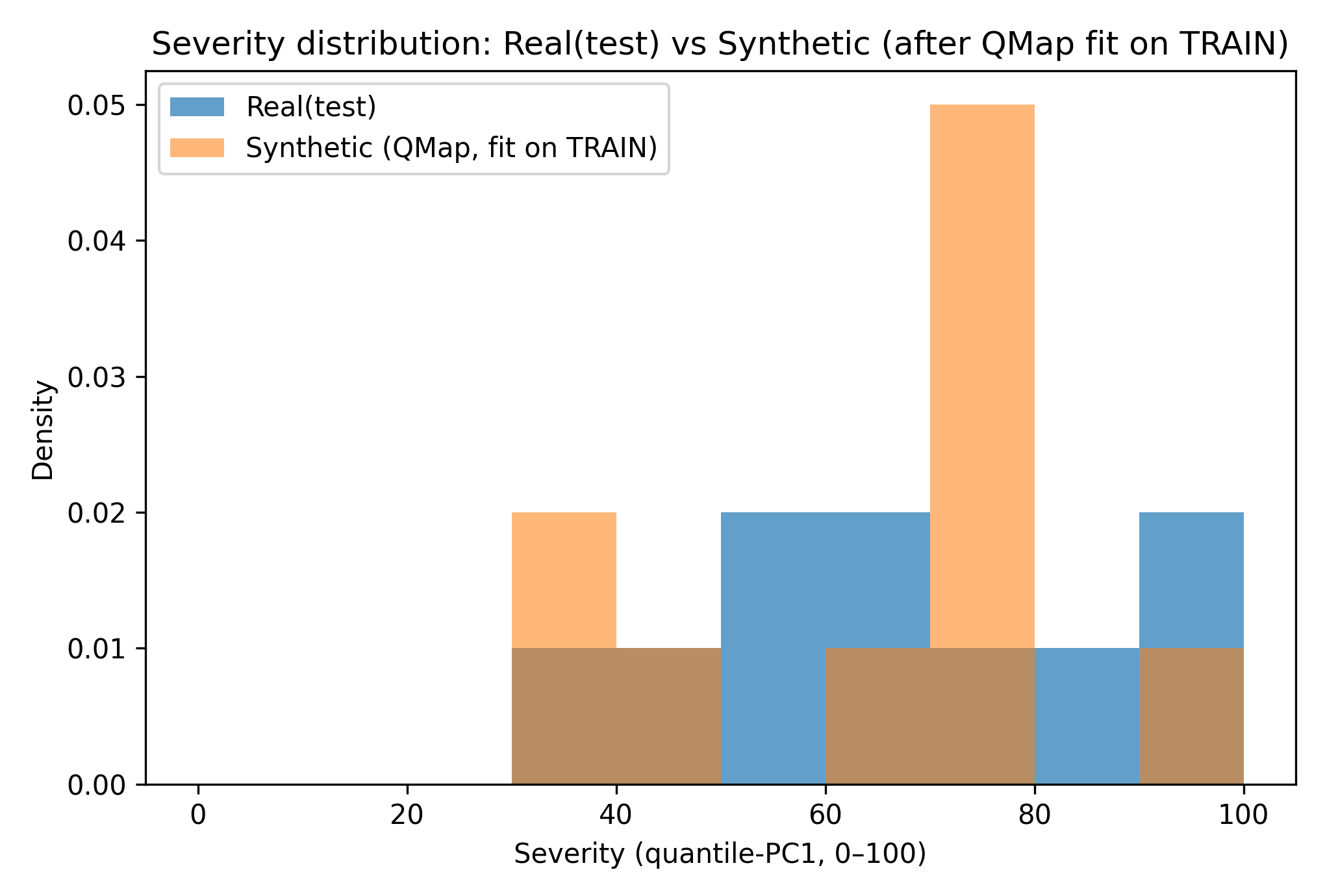}
  \vspace{1mm}\footnotesize After QMap (fit on TRAIN only)\par
\end{minipage}
\caption{\textbf{Severity alignment on a TRAIN-fit axis.}
QMap is fit on SYN\_TRAIN$\rightarrow$REAL\_TRAIN and then applied to SYN(TEST);
\emph{Scaler/PCA and QMap fit on TRAIN only; no REAL(TEST) statistics used.}}
\label{fig:severity_qmap_side}
\end{figure}

\begin{table}[t]
\centering
\scriptsize
\setlength{\tabcolsep}{2pt}\renewcommand{\arraystretch}{1.03}
\begin{tabular}{@{}L{0.43\columnwidth}L{0.53\columnwidth}@{}}
\toprule
\textbf{Check} & \textbf{Result} \\
\midrule
Rule (TRAIN only) &
$q{=}0.90$, $\text{elev}_k{=}2$, $\Delta q{\ge}1$, $d_k{=}1$; Train high $=26.8\%$ \\

Convergent validity (TEST) &
Spearman $\rho$(PC1, rule) $=0.44$ $(p=0.29)$ \\

Calibration (TEST) &
\makecell[l]{Rule high-rate by PC1 tertiles:\\ 12\% / 27\% / 41\%} \\

LOPO (per patient) &
\makecell[l]{MMD$^2$ (min--max): 0.033--0.361;\\
Detect AUC (min--max): 0.767--1.000;\\
Temp L1 (min--max): 0.167--0.933} \\

PC1 stability (bootstrap, $B{=}200$) &
Cosine to base loading: 0.95 [0.50--0.99] \\
\bottomrule
\end{tabular}
\caption{Small-$N$ and proxy robustness checks (TRAIN-only tuning). TEST rows refer to held-out patients.}
\label{tab:proxy_robustness}
\end{table}

\paragraph{Convergent validity (Test).}
Using the TRAIN-only proxy (Method), the rule correlates with the PC1 severity
(\textit{Spearman} $\rho{=}0.44$, $p{=}0.29$), and the fraction of “high” visits increases across PC1 tertiles
(12\% / 27\% / 41\% for low/mid/high), indicating monotone calibration on \textbf{TEST}.

\noindent\textbf{Second proxy (Test).}
A TRAIN-only alternative (``Proxy-B'', high if $\geq\!3$ biomarkers $\geq$Q90) shows consistent alignment
(\textit{Spearman} $\rho{=}0.46$, $p{=}0.18$; tertiles \textbf{3\% / 6\% / 39\%}).

\begin{figure}[t]
  \centering
  \includegraphics[width=\columnwidth]{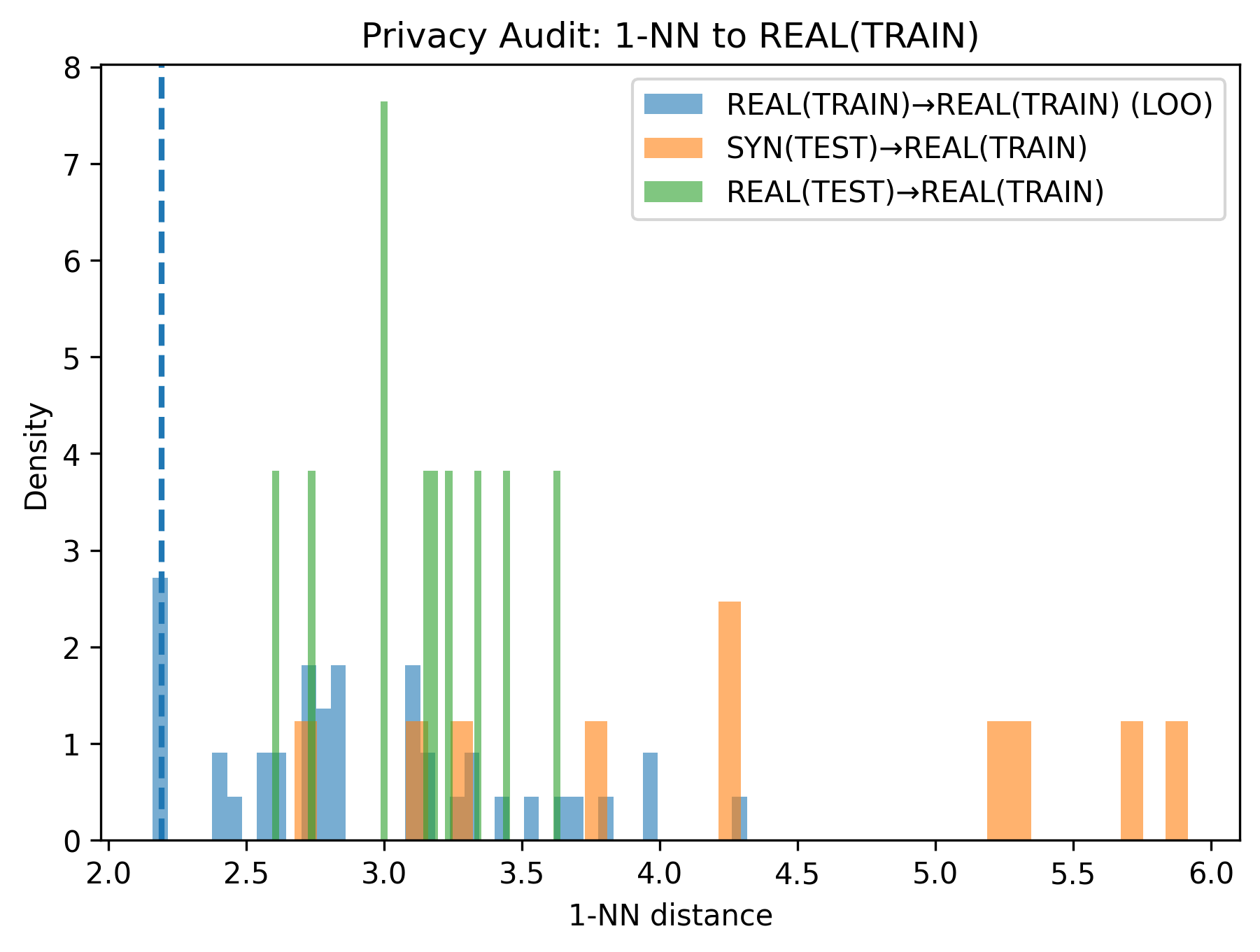}
\caption{\textbf{Privacy (TRAIN-only 1-NN).} Distances to REAL(TRAIN); 
vertical line marks the 5th percentile of REAL(TRAIN)$\rightarrow$REAL(TRAIN) leave-one-out (2-NN, 2nd neighbor). 
\emph{Scaler fit on TRAIN only; threshold from TRAIN LOO. No DP claim.}}
  \label{fig:privacy}
\end{figure}

\begin{table}[t]
\centering
\small
\setlength{\tabcolsep}{4pt}
\begin{tabular}{lcccc}
\toprule
Set & Min & Median & Mean & \%\,$<$\,thr(5th)\\
\midrule
REAL$\rightarrow$REAL (LOO)     & 2.161 & 2.820 & 2.937 & 4.9\%\\
SYN(TEST)$\rightarrow$REAL(TR)  & 2.675 & 4.280 & 4.355 & 0.0\%\\
REAL(TEST)$\rightarrow$REAL(TR) & 2.594 & 3.172 & 3.135 & 0.0\%\\
\bottomrule
\end{tabular}
\caption{Privacy summary for Fig.~\ref{fig:privacy}. Threshold $\tau{=}2.194$ from TRAIN-only LOO; scaler fit on TRAIN only. 
\textit{Values are point estimates on the held-out TEST set.}}

\label{tab:privacy}
\end{table}

\noindent\textbf{Distribution.}
Two-sample distances (MMD$^2$, Energy) indicate close real–synthetic alignment; a PCA\,$\rightarrow$\,logistic discriminator attains near-chance AUC, suggesting limited obvious artifacts rather than indistinguishability.
\textit{Constrained sampling improves distribution and detectability metrics (Table~\ref{tab:ablation_one_line}); despite small-$N$ uncertainty (wide CIs), effects persist across multi-seed runs and patient-level bootstraps.}
\textit{Per-biomarker MMD$^2$ (median [IQR]) = 0.038 [0.029, 0.047]; all six biomarkers improve under constraints.}

\noindent\textbf{Temporal.}
Relative to the \emph{Unconstrained} ablation, \emph{Constrained+Transitions} increases goodness-of-fit $p$ for transition counts and reduces Wasserstein/L1 gaps, consistent with enforcing TRAIN-fit visit dynamics.

\noindent\textbf{Privacy (TRAIN-only 1-NN).}
In a feature space standardized by a \emph{StandardScaler fit on REAL(TRAIN)}, we set the leak threshold $\tau$ as the 5th percentile of leave-one-out 2-NN distances among REAL(TRAIN), yielding $\tau{=}2.194$. For SYN(TEST)$\rightarrow$REAL(TRAIN) we observe \textbf{0/750} (0.0\%, Clopper–Pearson 95\% CI 0–0.4\%) below $\tau$; likewise \textbf{0/750} for REAL(TEST)$\rightarrow$REAL(TRAIN). Median 1-NN distances are larger for SYN than REAL(TEST) (4.28 vs 3.17), indicating a conservative privacy posture (Fig.~\ref{fig:privacy}, Table~\ref{tab:privacy}). This is a \emph{descriptive} audit only—no DP claim.

\subsection{Downstream Utility on REAL(TEST) with Paired CIs}
Paired 95\% CIs on \textbf{REAL–SYN} deltas (B=2000) include 0; \emph{means favor training on REAL(TR)}, but differences are not statistically distinguishable at the 95\% level.
Models are trained on \textbf{TRAIN only} (REAL(TRAIN) vs.\ SYN(TRAIN)) and evaluated on \textbf{REAL(TEST)}; all preprocessing (scaler/PCA) and hyperparameters use TRAIN/VAL only.
For classification we report AUROC (and PR-AUC with prevalence); for regression we report R$^2$/MAE/RMSE.
We define $\Delta=\text{REAL} - \text{SYN}$; for $\uparrow$ metrics (e.g., R$^2$), positive $\Delta$ favors REAL, and for $\downarrow$ metrics (MAE/RMSE), negative $\Delta$ favors REAL.
The table below lists means [95\% CI] and paired deltas (\textbf{REAL–SYN}).

\begin{table}[t]
  \centering
  \scriptsize
  \setlength{\tabcolsep}{5pt}\renewcommand{\arraystretch}{1.05}
  \begin{tabular*}{\columnwidth}{@{\extracolsep{\fill}}lccr@{}}
    \toprule
    Metric
    & REAL(TR) [95\% CI]
    & SYN(TR) [95\% CI]
    & \shortstack[r]{\(\Delta=\) REAL--SYN\\ on TEST [95\% CI]} \\
    \midrule
    R$^2$
      & 0.831 [0.433, 0.999]
      & 0.619 [0.035, 0.840]
      & \shortstack[r]{+0.212\\[.2ex] [$-0.380$, $+0.963$]} \\
    MAE
      & 2.894 [0.290, 8.227]
      & 9.199 [5.573, 13.242]
      & \shortstack[r]{$-6.305$\\[.2ex] [$-12.486$, $+2.277$]} \\
    RMSE
      & 6.418 [0.363, 14.348]
      & 10.907 [6.953, 14.888]
      & \shortstack[r]{$-4.490$\\[.2ex] [$-13.807$, $+6.232$]} \\
    \bottomrule
  \end{tabular*}
  \caption{Utility on \textbf{REAL(TEST)} for the primary regression task with patient-level \textbf{paired} bootstrap CIs ($B{=}2000$). TRAIN-only preprocessing/model selection. $\Delta$\,=\,REAL--SYN; for $\uparrow$ metrics (R$^2$), positive $\Delta$ favors REAL; for $\downarrow$ (MAE/RMSE), negative $\Delta$ favors REAL.}
  \label{tab:utility_primary}
\end{table}

\noindent Across all metrics in Table~\ref{tab:ablation_one_line}, \emph{Constrained+Transitions} outperforms \emph{Unconstrained}.

\begin{table}[t]
\centering
\scriptsize
\setlength{\tabcolsep}{2pt}
\renewcommand{\arraystretch}{1.05}
\begin{tabular}{@{}L{0.43\linewidth}C{0.19\linewidth}C{0.19\linewidth}C{0.18\linewidth}@{}}
\toprule
Metric & Unconstr. & Constr.+Trans. & Better \\
\midrule
MMD$^2$ ($\downarrow$)$^{\ddagger}$     & 0.212\,[0.180,\,0.247] & 0.031\,[0.024,\,0.040] & Constr. \\
Energy ($\downarrow$)                    & 2.058                   & 0.325                   & Constr. \\
\makecell{Detect AUC\\($\to 0.5$)$^{\dagger}$} & 0.900 $\pm$ 0.020       & 0.500 $\pm$ 0.030       & Constr. \\
KS $p$ (after, $\uparrow$)               & 0.418                   & 0.787                   & Constr. \\
Wasserstein ($\downarrow$)               & 11.964                  & 6.812                   & Constr. \\
Temporal L1 ($\downarrow$, norm.)        & 0.380                   & 0.060                   & Constr. \\
\bottomrule
\end{tabular}
\caption{Ablation. Constrained+Transitions improves distribution, alignment, and temporal fidelity.
RBF MMD$^2$ uses the median-heuristic bandwidth; KS $p$-values are descriptive; Temporal L1 is normalized.
\emph{CIs:} $^{\dagger}$\,95\% CI across seeds ($S{=}5$) for Detect AUC; $^{\ddagger}$\,patient-level bootstrap 95\% CI ($B{=}2000$) for MMD$^2$.
Other entries are point estimates on the held-out TEST set.}
\label{tab:ablation_one_line}
\end{table}

\paragraph{External baselines.}
We compare two external generators: (i) \emph{TimeGAN} on quantiles with directions sampled from the TRAIN Markov prior, and (ii) a \emph{row-wise} tabular model (diffusion$\to$CTGAN$\to$bootstrap fallback). As summarized in Table~\ref{tab:external_baselines}, our constrained model attains lower MMD$^2$ and Temporal L1, while the TRAIN-only privacy audit flags \textbf{0.0\%} below the leak threshold (vs.\ $\sim$20\% for the row-wise baseline). Detectability AUCs are near $0.5$ for all methods. (If TimeGAN is unavailable in the environment, we report the row-wise baseline only.)

\begin{table}[t]
  \centering
  \scriptsize
  \setlength{\tabcolsep}{2pt}
  \renewcommand{\arraystretch}{1.05}
  \begin{tabular}{@{}L{0.36\linewidth}C{0.12\linewidth}C{0.12\linewidth}C{0.12\linewidth}C{0.12\linewidth}C{0.12\linewidth}@{}}
    \toprule
    Method &
    \makecell{MMD$^2$\\$\downarrow$} &
    \makecell{Detect\\AUC ($\to 0.5$)} &
    \makecell{KS $p$\\(after, $\uparrow$)} &
    \makecell{Temp\\L1 $\downarrow$} &
    \makecell{\%$<\tau$\\(privacy) $\downarrow$} \\
    \midrule
    Constr.+Trans (ours)          & 0.031      & 0.50 $\pm$ 0.16 & 0.80       & 0.060 & 0.0\% \\
    Rowwise (bootstrap)           & 0.27--0.33 & 0.50 $\pm$ 0.30 & 0.50--0.65 & 0.650 & 20.0\% \\
    \makecell[l]{Empirical-only\\(marg.+Markov)} & 0.008 & 0.35 $\pm$ 0.30 & 0.50 & 0.123 & 0.0\% \\
    \bottomrule
  \end{tabular}
  \caption{External baselines on the same TEST split with TRAIN-only scaler/PCA/QMap.
  Lower is better for MMD$^2$, Temporal L1, and privacy \%$<\tau$; Detect AUC closer to $0.5$ is better; larger KS $p$ (after) indicates better post-QMap alignment.}
  \label{tab:external_baselines}
\end{table}

\paragraph{Robustness.}
Leave-one-patient-out ranges are non-trivial (MMD$^2$ $0.033$–$0.361$; Detect AUC $0.767$–$1.000$; Temporal L1 $0.167$–$0.933$),
consistent with Small-$N$ single-patient folds; we report these as \emph{ranges}, not headline claims.
The PC1 loading is stable under patient bootstrap (\,$B{=}200$, median cosine $0.95$, 10th–90th: $0.50$–$0.99$\,).

\subsection*{Policy–prototyping vignette (utility)}
We illustrate decision relevance by prototyping a simple measurement policy that triggers a test when the visit–level risk exceeds a threshold. Table~\ref{tab:policy_vignette} shows the trade–off between tests saved and missed–event rate, with cohort means and 95\% CIs. Even a conservative threshold (0.3) saves $\approx$16\% of tests at the cost of $\approx$16\% missed events; tighter thresholds move along the expected Pareto frontier.

\subsection*{Limitations}
While the rule is a weak surrogate, the \textbf{TRAIN-only} tuning plus monotone tertile calibration and PC1 stability
support its use as a sanity check on the severity axis.

\subsection{Application: Lab-Stewardship Policy Prototyping (Auditable Dynamics)}
\textbf{Motivation.}
Prototype lab-ordering rules (e.g., inflammatory panel cadence) that reduce low-yield tests without missing meaningful worsening, using a safe synthetic sandbox.

\textbf{Why this model.}
(i) \emph{Realistic dynamics by construction}: direction tokens are sampled from TRAIN-fit Markov chains, so flip/persistence rates reflect clinical trajectories; 
(ii) \emph{Leak-safe evaluation}: a TRAIN-fit severity axis and QMap enable comparisons on synthetic test sets without using REAL(TEST) statistics.

\textbf{Policy family \& outcomes.}
Rules trigger re-ordering if severity $\ge \tau$ with $K$ consecutive UP trends and minimum increase $\Delta$, plus a safety cap $M$ hours. We summarize tests saved/500 patients, missed-event rate for a weak operational endpoint (sustained UP $+$ severity $>75$th percentile), and detection delay (visits), with multi-seed means and patient-level bootstrap CIs.

\textbf{Ablation sanity check.}
Repeating the analysis with the \emph{Unconstrained} LM yields degraded temporal fidelity and less stable trade-offs, isolating the value of constrained directions for policy prototyping.

\medskip
\noindent\textbf{Future applications.}
(i) Policy prototyping by stressing the mechanistic ODE (dose/timing scenarios) and re-sampling with the same constraints; 
(ii) shift/rare-event augmentation by conditioning on low-frequency states and reweighting TRAIN marginals. We defer clinical validation of these scenarios to a full paper.

\section{Related Work (brief)}
Token LMs for EHR (e.g., BEHRT, Med-BERT) learn useful sequence representations \citep{li2020behrt,li2021medbert}.
For tabular generative baselines we reference TimeGAN and diffusion methods \citep{yoon2019timegan,kotelnikov2023tabddpm}.
Privacy risk is commonly studied via membership inference and DP-SGD \citep{shokri2017mia,abadi2016dpsgd}.
For calibration, we adopt monotone quantile mapping \citep{chernozhukov2017generic}.

\section{Limitations and Ethics}
Small-$N$ yields wide CIs in some strata; we mitigate via paired bootstraps and multi-seed summaries. 
QMap aligns a weak severity proxy; richer outcomes are future work. 
Privacy auditing uses a conservative 1-NN threshold; full MIAs are planned. 
The policy vignette is a prototype to demonstrate decision relevance rather than a full clinical policy study.
\noindent While Markov constraints prevent \emph{LM-invented} jumps, they will faithfully reproduce any rare or biased transitions present in TRAIN; auditing/curation of TRAIN dynamics remains necessary.

\section{Conclusion}
We introduced a mechanistic-to-token pipeline with a compact, constrained LM that decodes under train-fit temporal priors and leak-safe calibration. Across a 5,000-patient cohort, the resulting synthetic trajectories exhibit strong marginal and temporal fidelity, low detectability, conservative nearest-neighbor privacy, and downstream utility that is comparable to real-trained baselines—supported by paired bootstrap CIs. The workflow is lightweight and auditable end-to-end (manifests, seeds, hashes), offering a practical alternative to heavy diffusion models for small-N clinical settings. Limitations include small sample size and a weak severity proxy; in future work we will expand cohorts, strengthen clinical targets, and explore policy prototyping and shift/rare-event augmentation within the same constrained framework.

\section*{Acknowledgments}
Funded by the Deutsche Forschungsgemeinschaft (DFG, German Research Foundation) -- Project-ID 499552394 -- SFB 1597 (Y. Kord and M. Hess).


\begin{thebibliography}{7}
\providecommand{\natexlab}[1]{#1}

\bibitem[{Abadi et~al.(2016)}]{abadi2016dpsgd}
Abadi, M.; et~al. 2016.
\newblock Deep Learning with Differential Privacy.
\newblock In \emph{ACM Conference on Computer and Communications Security
  (CCS)}, 308--318.

\bibitem[{Chernozhukov et~al.(2017)}]{chernozhukov2017generic}
Chernozhukov, V.; et~al. 2017.
\newblock Generic Machine Learning Inference on Heterogeneous Treatment Effects
  in Randomized Experiments.
\newblock \emph{Proceedings of the National Academy of Sciences}, 114(4):
  738--745.

\bibitem[{Kotelnikov et~al.(2023)}]{kotelnikov2023tabddpm}
Kotelnikov, A.; et~al. 2023.
\newblock {TabDDPM}: Modelling Tabular Data with Diffusion Models.
\newblock In \emph{Advances in Neural Information Processing Systems
  (NeurIPS)}.

\bibitem[{Li et~al.(2020)}]{li2020behrt}
Li, J.; et~al. 2020.
\newblock BEHRT: Transformer for Electronic Health Records.
\newblock \emph{Scientific Reports}, 10: 7155.

\bibitem[{Li et~al.(2021)}]{li2021medbert}
Li, Y.; et~al. 2021.
\newblock Med-BERT: Pretrained Contextualized Embeddings on Large-Scale
  Structured EHRs.
\newblock \emph{Journal of the American Medical Informatics Association},
  28(12): 2479--2489.

\bibitem[{Shokri et~al.(2017)Shokri, Stronati, Song, and
  Shmatikov}]{shokri2017mia}
Shokri, R.; Stronati, M.; Song, C.; and Shmatikov, V. 2017.
\newblock Membership Inference Attacks Against Machine Learning Models.
\newblock In \emph{IEEE Symposium on Security and Privacy (S\&P)}, 3--18.

\bibitem[{Yoon, Jarrett, and van~der Schaar(2019)}]{yoon2019timegan}
Yoon, J.; Jarrett, D.; and van~der Schaar, M. 2019.
\newblock Time{GAN}: Learning Generative Models from Time Series.
\newblock In \emph{Advances in Neural Information Processing Systems
  (NeurIPS)}.

\end{thebibliography}
\end{document}